\documentclass[conference]{IEEEtran}
\IEEEoverridecommandlockouts
\usepackage{times}
\usepackage{graphicx}
\usepackage{amsmath}
\usepackage{amssymb}
\usepackage{etoolbox}
\usepackage{amsthm}
\usepackage{bm}
\usepackage{xspace}
\usepackage{standalone}
\usepackage{pgfplots}
\usepackage{filecontents}
\usepackage{tikz}
\usepackage[T1]{fontenc}
\usepackage{subcaption}
\usepackage{mathtools}
\usepackage{enumitem}
\usepackage[ruled,vlined,noend]{algorithm2e}
\usepackage{amssymb,amsmath,amsthm}

\newtheorem{lemma}{Lemma}

\providecommand{\fref}[1]{Fig.~\ref{#1}} 

\providecommand{\Language}[1]{\ensuremath{\mathcal{L}(#1)}}

\providecommand{\gobble}[1]{}

\providecommand{\term}{\ensuremath{\mathrm{term}}}
\providecommand{\goal}{\ensuremath{\mathrm{goal}}}

\newrobustcmd*{\mysquare}[1]{\tikz{\filldraw[draw=black,fill=#1] (0,0)
rectangle (0.2cm,0.2cm);}}

\newrobustcmd*{\mycircle}[1]{\tikz{\filldraw[draw=black,fill=#1] (0,0) circle [radius=0.1cm];}}

\newrobustcmd*{\mytriangle}[1]{\tikz{\filldraw[draw=black,fill=#1] (0,0) --
(0.2cm,0) -- (0.1cm,0.2cm);}}

\definecolor{amethyst}{rgb}{0.6, 0.4, 0.8}
\definecolor{gre}{rgb}{0.0, 0.5, 0.0} 
\definecolor{gra}{rgb}{0.8, 0.8, 0.85} \definecolor{re}{rgb}{1.0, 0.5, 0.5}

\definecolor{setformula}{rgb}{0.0, 0.0, 0.0}
\definecolor{equivrel}{rgb}{0.0, 0.0, 0.0}

\definecolor{darkblue}{rgb}{0.15,0.09,0.3}

\definecolor{formula}{rgb}{0.08,0.35,0.08}
\newcommand\form[1]{\textcolor{formula}{\bm{#1}}}
\definecolor{varcolour}{rgb}{0.0,0.0,0.0}

\newrobustcmd*{\true}{{\sl True}\xspace}
\newrobustcmd*{\false}{{\sl False}\xspace}

\DeclareFontFamily{OT1}{pzc}{}
\DeclareFontShape{OT1}{pzc}{m}{it}{<-> s * [1.10] pzcmi7t}{}
\DeclareMathAlphabet{\mathpzc}{OT1}{pzc}{m}{it}

\newcounter{tecounter}
\usepackage[noend]{algorithmic}
\algsetup{linenosize=\tiny}

\providecommand{\sde}[1]{\ensuremath{\operatorname{\textsc{Sde}}({#1})}}

\newcommand{\defeq}{\ensuremath{\coloneqq}}
\providecommand{\True}{\ensuremath{\operatorname{{True}}}\xspace}

\providecommand{\reachedv}[2]{\textcolor{setformula}{\ensuremath{\mathcal{V}^{#1}_{#2}}}}

\providecommand{\exactreachings}[2]{\textcolor{setformula}{\ensuremath{\mathbb{S}}^{#1}_{#2}}}

\providecommand{\compatablew}[2]{\textcolor{setformula}{\ensuremath{\mathcal{W}}^{#1}_{#2}}}

\usepackage[numbers]{natbib}
\usepackage{multicol}
\usepackage{tabularx}
\usepackage[bookmarks=true]{hyperref}
\usepackage{xcolor}
\hypersetup{
  colorlinks,
  linkcolor={red!50!black},
  citecolor={blue!50!black},
  urlcolor={blue!80!black}
  }

\def\BibTeX{{\rm B\kern-.05em{\sc i\kern-.025em b}\kern-.08em
    T\kern-.1667em\lower.7ex\hbox{E}\kern-.125emX}}
\begin{document}

\title{What does my knowing your plans tell me?
}

\newcommand\unsure[1]{\textcolor{brown}{{#1}}}

\author{
\IEEEauthorblockN{Yulin Zhang}
\IEEEauthorblockA{\textit{Dept. of Comp. Sci. \& Engr.} \\
\textit{Texas A\&M University}\\
College Station, Texas, USA\\
yulinzhang@tamu.edu}
\and
\IEEEauthorblockN{Dylan A. Shell}
\IEEEauthorblockA{\textit{Dept. of Comp. Sci. \& Engr.} \\
\textit{Texas A\&M University}\\
College Station, Texas, USA\\
dshell@tamu.edu}
\and
\IEEEauthorblockN{Jason M. O'Kane}
\IEEEauthorblockA{\textit{Dept. of Comp. Sci. \& Engr.} \\
\textit{University of South Carolina}\\
Columbia, South Carolina, USA\\
jokane@cse.sc.edu}
}

\maketitle

\begin{abstract}
For robots acting in the presence of observers, we examine the information that
is divulged if the observer is party to the robot's plan.
Privacy constraints are specified as the
stipulations on what can be inferred during plan execution. 
We imagine a case in which the robot's plan is divulged beforehand, so
that the observer can use this {\em a priori} information 
along with the disclosed executions. 
The divulged plan, which can be represented by a procrustean graph, is shown to
undermine privacy precisely to the extent that it can eliminate
action-observation sequences that will never appear in the plan. 
Future work will consider how the divulged plan might be sought as the output of a planning procedure.
\end{abstract}

\section{Introduction}
Autonomous robots are beginning to be part of our everyday lives.  Robots may
need to collect information to function properly, but this information can be
sensitive if leaked. In the future,
robots will not only need to ensure physical safety for humans in shared
workspaces, but also to guarantee their information security. 
But information leakage can occur in a variety of ways, including
through logged data, robot's status display, actions, or, as we examine,
through provision of prior information about a robot's plan. 

Established algorithmic approaches for the design and implementation of
planners may succeed at
selecting actions to accomplish goals, but they fail to consider what
information is divulged along the way. While several models for privacy exist,
they have tended to be either abstract definitions applicable to data rather
than an agent operating autonomously in the world (such as
encryption~\cite{menezes96crypto}, data synthesis~\cite{rubin93synth}, 
anonymization~\cite{Dwork2008}, or opacity~\cite{jacob16opacity} mechanisms) or are focussed on a particular
robotic scenario (such as robot division of labor~\cite{prorok2016macroscopic}
or tracking~\cite{OKa08, zhang18complete}). 

\begin{figure}[ht!]
\vspace*{-4pt}
\centering
\includegraphics[scale=0.38]{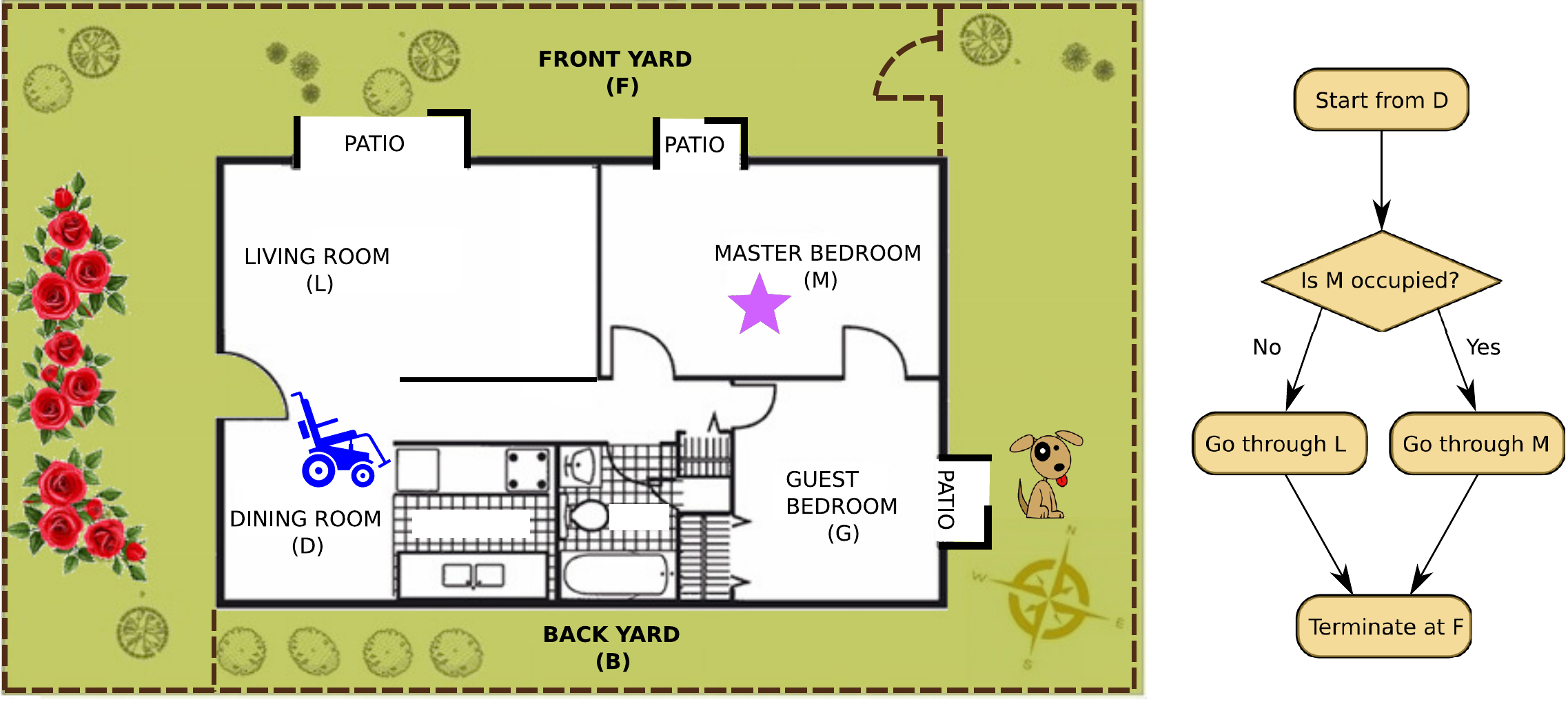}
\caption{An autonomous wheelchair navigates in a home.
A plan, on the right, generates actions that depend on perception of
the pink star (denoting that the bedroom is occupied). 
\label{fig:wheelchair}}
\end{figure}

Figure \ref{fig:wheelchair} illustrates a scenario 
where the information divulged is subtle and important.  It considers an
autonomous wheelchair that helps a patient who has difficulty navigating by
himself. The user controls the wheelchair by giving voice commands: once the
user states a destination, the wheelchair navigates there autonomously.  While
moving through the house, the wheelchair should avoid entering any
occupied bedrooms, making use of information from motion sensors installed
inside each bedroom.  We are interested in stipulating the information divulged
during the plan execution:
\begin{description}
\item  \underline{Positive disclosure of information}: 
A therapist monitors the user, ensuring that he adheres to his daily regimen
of activity, including getting some fresh air everyday (by visiting
the front yard or back yard).
\item  \underline{Ne}g\underline{ative disclosure of information}: However, if
there is a guest in one of the bedrooms, the user does not want to disclose the
guest's location. 

\end{description}

Actions, observations, and other information (such as the robot's planned
motion) may need to be divulged to satisfy the first (positive) stipulation.
The challenge is to satisfy both stipulations simultaneously.  Suppose the
robot executes the plan shown in the right of \fref{fig:wheelchair}, and that
this plan is public knowledge.  If, as it moves about, the robot's observations (or actions) are
disclosed to an observer, then we know that the robot will attempt to see if
$M$ is occupied.  Hence, on some executions, a third party, knowing
there is a guest, would be able to infer that they're in the master bedroom.

This paper examines in detail how divulging the plan, as above, provides
information that permits one to draw inferences.
In particular, we are interested in how this plan information 
might cause privacy violations.  As we will see, the divulged plan need not
be the same as the plan being executed, but they must agree in a certain way.
In our future work, we hope to answer the question of how to find pairs of
plans (one be to executed and one to divulged), where there is some \emph{gap}
between the two, so that information stipulations are always satisfied.

\begin{figure*}[t!]
\centering
\includegraphics[scale=0.85]{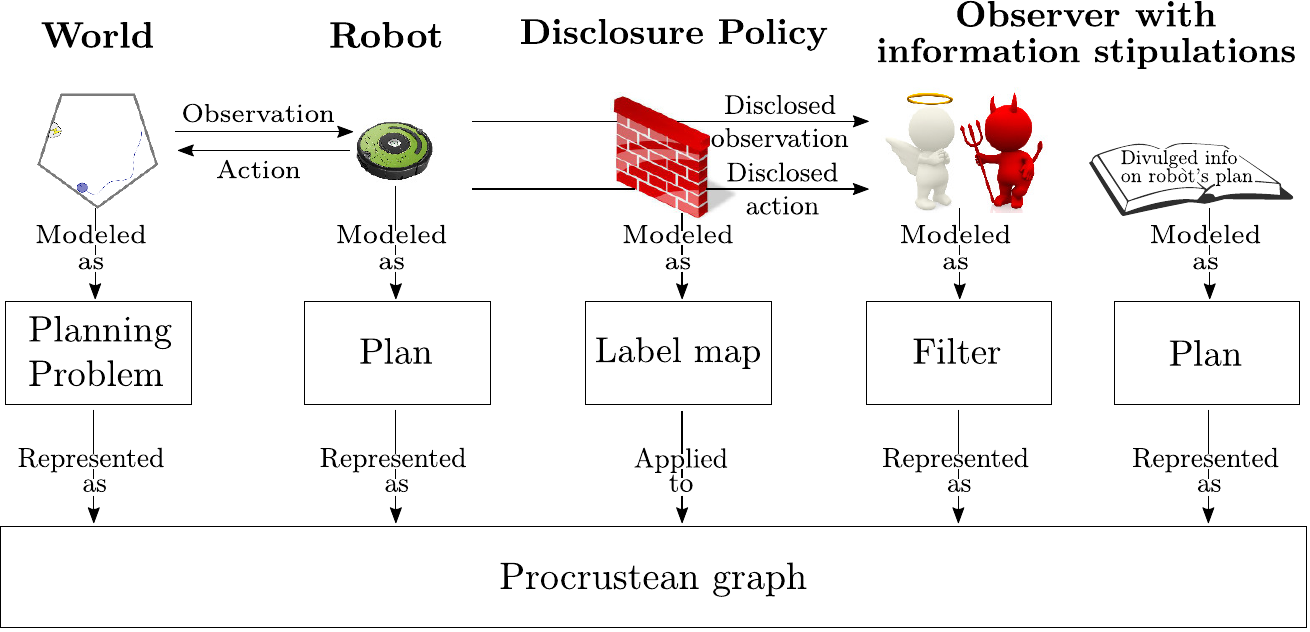}
\caption{An overview of the setting: the robot is modeled abstractly as realizing
a plan to achieve some goal in the world and the third party observer as a
filter with divulged plan as its prior knowledge.  All four, the world, the plan, the filter, and the divulged plan have concrete
representations as p-graphs.
\label{fig:modeloverview}}
\end{figure*}

\vspace*{-8pt}
\section{Problem Description}

In this problem, there are three entities: a \emph{world}, a \emph{robot}, and
an \emph{observer}. As shown in \fref{fig:modeloverview}, the robot interacts
with the world by taking observations from the world as input, and outputting
an action to influence the world state. This interaction generates a stream of
actions and observations, which may be perceived by the observer, though
potentially only in partial or diminished form.  We model the stream as
passing through a function which, via conflation, turns the stream generated by
the world--robot interaction into one perceived by the observer, the disclosed action-observation stream. As a
consequence of real-world imperfections (possible omission, corruption, or
degradation) or due to explicit design, the observer,  thus, may receive less
information. For this reason, the function is viewed as a sort of barrier, and
we term it an \emph{information disclosure policy}.  

The observer is assumed to be unable to take actions to interact with the world directly---a model
that is plausible if the observer is remote, say a person 
or service on the other side of a camera or other Internet of Things device. 
Given its perception of the interaction, 
the observer estimates the plausible action-observation streams, consistent with the disclosed action-observation
stream. This estimate can be made `tighter' by leveraging prior knowledge
about the robot's plan.  The observer's estimate is in terms of world states,
so the notion of tightness is just a subset relation.  In this paper, we will
introduce stipulations on these estimated world states and our main
contribution will be in examining how the divulged plan could affect the
satisfaction of these stipulations.

\subsection{Representation} 

To formalize such problem, we represent these
elements with p-graph formalism and label map \cite{saberifar18pgraph}. The
world is formalized as a planning problem $(W, V_{\goal})$, where $W$ is a
p-graph in state-determined form (see definition of state-determined in
\cite[Def.~3.7]{saberifar18pgraph}) and $V_{\goal}$ is the set of goal states. The robot
is modeled as a plan $(P, V_{\term})$, where $P$ is a p-graph and $V_{\term}$
specifies the set of plan states where the plan could terminate. The plan
solves the planning problem when the plan can always safely terminate at the
goal region in finite number of steps (see definition of solves in
\cite[Def.~6.3]{saberifar18pgraph}). The information disclosure policy is represented by
a label map $h$, which maps from the actions and observations from $W$ and $D$
to an image space $X$. The observer is modeled as a tuple $(I, D)$, where $I$
is a filter represented by a p-graph with edge labels from $X$, $D$ is the
p-graph representing the divulged plan with actions and observations labeled in
the domain of $h$. 
The plan in $D$ might be less specific than the actual plan $P$, representing
`diluted' knowledge of the plan; to capture this, we require that all possible
action-observation sequences (called executions for short) in $D$ should be a
superset of those in $P$, denoted as $\Language{D}\supseteq \Language{P}$ (the
set of executions is called the language,
see~\cite[Def.~3.5]{saberifar18pgraph}, hence the symbol $\Language{\cdot}$).

\subsection{The observer's estimation of world states}

Given any set of filter states $B$ from filter $I$, the observer obtains an
estimate of the executions that should've occurred to reach $B$, through a
combination of the following sources of information ~\cite[Def.~13]{zhang18planning}:

\begin{enumerate}
\item The observer can ask: What are all the possible executions, each of which has its image, reaching exactly $B$ in the filter? The set of executions reaching exactly $B$ is represented as $\exactreachings{I}{B}$. The preimages of $\exactreachings{I}{B}$, which we denote as $h^{-1}[\exactreachings{I}{B}]$, are the executions which are responsible for arriving at $B$ in $I$.
\item The observer can narrow down the estimated executions to the ones that only appear in the divulged plan $D$. The set of all executions in $D$ are represented by its language~$\Language{D}$.
\item Finally, the estimated executions can be further refined by considering those that appear in the world, i.e., $\Language{W}$.
\end{enumerate}

Hence, $h^{-1}[\exactreachings{I}{B}]\cap \Language{W}\cap \Language{D}$
represents a tight estimation of the executions that may happen. This allows us
to find the estimated world states, defined as $\compatablew{D}{B}$, by making
a tensor product $T$ of graph $W$, $D$ and $h^{-1}\langle I\rangle$, where
$h^{-1}\langle I\rangle$ is obtained by replacing each action or observation
$\ell$ with its preimage $h^{-1}(\ell)$ on the edges of the p-graph $I$. For
any vertex $(w, d, i)$ from the product graph $T$, we have:
\[\compatablew{D}{B}=\compatablew{D}{B}\cup \{w\}, {\rm if}~ i\in B.\]

\subsection{Information stipulations on the estimated world states}

Information stipulations are written as propositional formulas on 
estimated world states $\compatablew{D}{B}$. Firstly, we will define a symbol
$\mathpzc{w}$ for each world state $w$ in $W$. Then we can use connectives
$\neg$, $\land$, $\lor$ to form composite expressions $\form{\Phi}$ that
stipulate the estimated world states involving these symbols. The propositional
formulas can be evaluated based on the following definition:
\[\mathpzc{w}=\True\quad\text{if and only if} \quad w\in \compatablew{D}{B}.\]

With all the elements defined above, we are able to check whether the
stipulation $\form{\Phi}$ is satisfied on every estimate $\compatablew{D}{B}$,
given the world graph $W$, information disclosure policy $h$, and the observer
$(I, D)$.

\section{The observer's prior knowledge of the robot's plan}

The divulged plan $D$ is known by the observer prior to the robot's monitoring
of the disclosed action-observation stream. Depending on how much the observer
knows, there are four possibilities, from most-to least-informed:

\renewcommand{\theenumi}{\Roman{enumi}}%
\begin{enumerate}
\item The observer knows the exact plan $P$ to be executed.\label{item:exactplan}
\item The plan to be executed can be hidden among a (non-empty) finite set of plans
$\{P_1,P_2, \dots, P_n\}$.\label{item:setplan}
\item The observer may only know that the robot is executing \emph{some} plan, that is, the robot is goal directed and aims to achieve some state in $V_{\goal}$. \label{item:someplan}
\item The observer knows nothing about the robot's execution other than that it is on $W$. \label{item:wanderingrobot}
\end{enumerate}
A p-graph exists whose language expresses knowledge for
each of these cases:
\begin{description}[style=unboxed,leftmargin=0cm]
\item Case~\ref{item:exactplan}.\quad
When $D=P$, the interpretation is straightforward: the observer
tracks the states of the plan  given the stream of observations (as best as
possible, as the operation is under $h$). 
\item Case~\ref{item:setplan}.\quad
If instead a set of plans $\{P_1,P_2, \dots, P_n\}$ is given, we
must construct a single p-graph, $D$, so that $\Language{D} = \Language{P_1}
\cup \dots  \cup \Language{P_n}$. This is achieved via the union of p-graphs $D
= P_1 \uplus P_2 \uplus \dots \uplus P_n$, cf.~\cite[Def.~3.6,  pg.~18]{saberifar18pgraph}.
\item Case~\ref{item:someplan}.\quad
If the robot is known only to be executing some plan, we must consider the set of
all plans, ${P^{\infty} \defeq \{P_1,P_2, P_3, \dots, \}}$.
As the notation hints,  there can be an infinite number of such
plans, so the approach of unioning plans won't work.
Fortunately, another structure, $P^{*}$, exists such that $\Language{D}=\Language{P^{*}}=\Language{P^{\infty}}$, which will be proved afterwards.
Here $P^{*}$, a finite p-graph, is called the \emph{plan closure}. 
\item Case~\ref{item:wanderingrobot}.\quad
When taking $D=W$ the executions are,
again, intersected with $\Language{D}$ but as they already came from
$\Language{W}$, this shows why the observer is the least informed in the
hierarchy. 
\end{description}

Next, we will show the construction of the plan closure $P^{*}$ and prove that $\Language{P^{*}}=\Language{P^{\infty}}$.


To start, we describe construction of $P^{*}$. The initial step is to convert $W$ to
its state-determined form $W'=\sde{W}$ (this is an operation described in 
\cite[Algorithm~2, pg.~30]{saberifar18pgraph}).  Then, to decide whether a vertex in $W'$ exists in
some plan, we iteratively color each vertex green, red, or gray. Being colored
green means that the vertex exists in some plan, red means that the vertex does
not exist in any plan, and gray indicates that its status has yet to be
decided.  To start with, we initially color the goal vertices green, and
non-goal leaf vertices (with no edges to other vertices) red.  Using the
iconography of \cite{saberifar18pgraph}, we show action vertices as squares and
observation vertices as circles.  Then gray vertices of each type change their
color by iterating the following steps:
\begin{itemize}
\item \mysquare{gra} $\rightarrow$ \mysquare{gre}: $\exists$ some action $a$ reaching $\mycircle{gre}$, which is not an initial state.
\item \mysquare{gra} $\rightarrow$ \mysquare{re}: $\forall$ action $a$ reaching $\mycircle{re}$.
\item \mycircle{gra} $\rightarrow$ \mycircle{gre}: $\forall$ observation $o$ reaching $\mysquare{gre}$, which is not an initial state.
\item \mycircle{gra} $\rightarrow$ \mycircle{re}: $\exists$ some observation $o$ reaching $\mysquare{re}$.
\end{itemize}
The iteration ends when no vertex changes its color. 
The subgraph that consisting of only green vertices and their corresponding edges
is $P^{*}$. And $P^{*}$ then contains only the vertices that exist in some
plan leading to the goal states. For further detail of this algorithm for building $P^{*}$,
we refer the reader to Algorithm~\ref{alg:pstar}.

\begin{figure}
\vspace*{-20pt}
 	\centering
 	\includegraphics[width=0.8\linewidth]{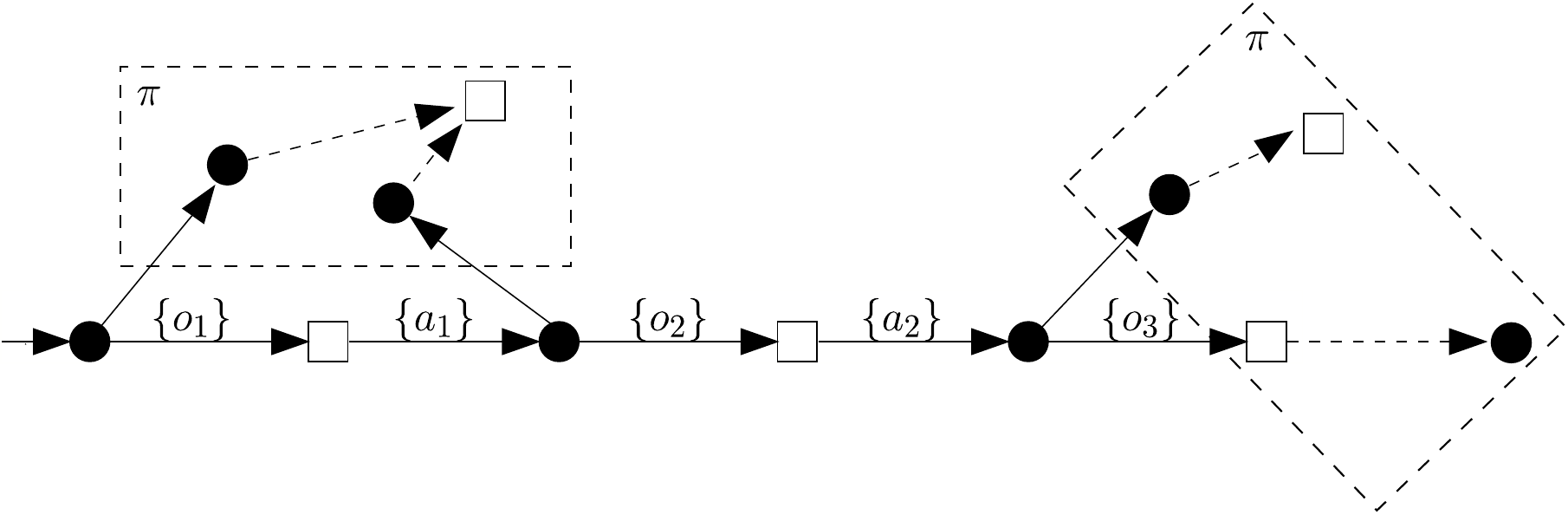}
        \caption{The construction of a plan generating execution~$s$ using~$\pi$,
        computed as part of Algorithm~\ref{alg:pstar}.
        } \label{fig:s-pi}
 \vspace*{-15pt}
\end{figure}

Next, we prove that the $P^{*}$ constructed from this procedure
has the same language as $P^{\infty}$. The proof shows that
any green vertex is on some plan, by showing that we 
we can construct a plan $\pi$,
that will lead to a goal state within a finite number of
steps form any such vertex.

\begin{lemma}
\label{lem:pinfinity}
\Language{P^{*}}=\Language{P^{\infty}}.
\end{lemma}
\vspace*{-8pt}
\begin{proof}
$\supseteq$: For any $s=s_0s_1s_2\dots s_k\in \Language{P^{\infty}}$,
according to the definition of $P^{\infty}$, $s$ is in the execution of some
plan $P'$. Though $s_k$ may not be a goal, using $P'$, $s$ can be extended:
$\exists s'=s_0s_1\dots s_k t_0 t_1\dots t_n\in \Language{P'}$, $k > 0,n\geq 0$
to reach an element of $V_\goal$. Then $\reachedv{P'}{s'}$ comprises 
vertices associated with those in $W'$ marked green in $V'_\goal$. And, tracing
the execution $s'$ on $P'$ backwards on $W'$, we find every vertex green back
to a start vertex. But this means they
are in $P^*$, and hence $s' \in \Language{P^{*}}$, means $s \in
\Language{P^{*}}$ as well.

$\subseteq$: For any execution $s=s_0s_1s_2\dots s_k\in \Language{P^{*}}$, $s$
reaches $V'_{\goal}$, or $s$ is a prefix of some execution reaching $V'_{\goal}$
in $W'$.  We show that there is a plan that can produce $s$. The execution $s$
does not include enough information to describe a plan because: (1)~it may not
reach $V'_{\goal}$ itself, and (2)~it gives an action after some observation
that was revealed, but not every possible observation.  To address this
shortfall, we will capture some additional information during the construction
of $P^{*}$, which we save in $\pi$. This provides an action that makes some
progress, for states that can result from other observations.  Now, using $s$
as a skeleton, construct plan where once a transition outside of $s$ occurs,
either owing to an unaccounted-for observation or having reached the end of
$s$, the plan reverts to using the actions that $\pi$ prescribes. 
(See \fref{fig:s-pi} for a visual example.) This is always possible
because states arrived at in $W'$ under $s$ are green. This implies that all
states in $W$ are also assured to reach a goal states.
The resulting plan can produce $s$, so some plan produces $s$, hence $s \in \Language{P^{\infty}}$.
\end{proof}

\begin{algorithm}[ht!]
\caption{$P^*${\sc Construction}$(W,V_{\goal})$}
\label{alg:pstar}
\begin{algorithmic} 
	\STATE Initialize queues $\rm red$, $\rm green$, $\rm gray$ as empty
	\STATE $W'\gets \sde{W}$, and initialize $V'_{\goal}$ as the associated vertices of $V_{\goal}$
	\STATE Initialize plan $\pi$ as empty
	\FOR{$v\in V(W')$}
		\IF{$v\in V'_{\goal}$}
			\STATE $\rm green$.append($v$)
		\ELSIF{$v$ has no edges to other vertices}
			\STATE $\rm red$.append($v$)
		\ELSE
			\STATE $\rm gray$.append($v$)
		\ENDIF
	\ENDFOR
	\STATE $Q$.extend(InNeighbor(${\rm red} \cup {\rm green}$)$\backslash ({\rm red} \cup {\rm green})$)
	\WHILE{$Q$ not empty}
		\STATE $v\gets Q$.pop
		\STATE ${\rm flag}\gets$\true
		\IF{$v$ is a \mycircle{gra}}
			\IF{one of its outgoing neighbors is \mysquare{re}}
				\STATE $\rm red$.append($v$)
			\ELSIF{all of its outgoing neighbors are \mysquare{gre}}
				\STATE $\rm green$.append($v$)
			\ELSE
				\STATE ${\rm flag}\gets$\false
			\ENDIF
		\ELSIF{$v$ is a \mysquare{gra}}
			\IF{one of its outgoing neighbors under label $a$ is \mycircle{gre}}
				\STATE $\rm green$.append($v$) and $\pi[v]=a$
			\ELSIF{all of its outgoing neighbors are \mycircle{re}}
				\STATE $\rm red$.append($v$)
			\ELSE
				\STATE ${\rm flag}\gets$\false
			\ENDIF
		\ENDIF
		\IF {$\rm flag$}
			\STATE $Q$.extend(InNeighbor($v$)$\backslash\{{\rm red} \cup {\rm green}\}$)
		\ENDIF
	\ENDWHILE
	\STATE $P^{*}\gets$ subgraph($W'$, ${\rm green}$)
	\RETURN $P^{*}$ (and also $\pi$, if desired)
\end{algorithmic}
\end{algorithm}

Thus, one may use $D = P^*$, for Case~\ref{item:someplan}.

\section{Experimental results}
We implemented the algorithms with Python, and execute them on a OSX laptop
with a 2.4 GHz Intel Core i5 processor. To experiment, we constructed a p-graph
representing the world in \fref{fig:wheelchair} with $12$ states, and the plan
with $8$ states. All the experiments are finished within $1$ second. The
information disclosure policy maps all actions to the same image, but
observations to different images. As we anticipated, the stipulations are
violated when the exact plan is divulged. But we can satisfy the stipulations
by disclosing less information, such as $D=W$.

\section{Summary and future work}

We examine the planning problem and the information divulged within the
framework of procrustean graphs.  In particular, the divulged plan can be
treated uniformly in this way, despite representing four distinct cases.  The
model was evaluated, showing that divulged plan information can prove to be a
critical element in protecting the privacy of an individual.  In the future, we
aim to automate the search for plans: given $P$ to be executed, find a $D$ to
be divulged, where $\Language{D}\supsetneq \Language{P}$, such that the privacy
stipulations are always satisfied.

\section*{Acknowledgements}

This work was supported by the NSF through awards 
\href{http://nsf.gov/awardsearch/showAward?AWD_ID=1453652}{IIS-1453652},
\href{http://nsf.gov/awardsearch/showAward?AWD_ID=1527436}{IIS-1527436},
and 
\href{http://nsf.gov/awardsearch/showAward?AWD_ID=1526862}{IIS-1526862}.
We thank the anonymous reviewers for their time and valuable comments.

\IEEEpeerreviewmaketitle

\bibliographystyle{IEEEtran}
\bibliography{mybib}

\begin{thebibliography}{1}
\providecommand{\url}[1]{#1}
\csname url@samestyle\endcsname
\providecommand{\newblock}{\relax}
\providecommand{\bibinfo}[2]{#2}
\providecommand{\BIBentrySTDinterwordspacing}{\spaceskip=0pt\relax}
\providecommand{\BIBentryALTinterwordstretchfactor}{4}
\providecommand{\BIBentryALTinterwordspacing}{\spaceskip=\fontdimen2\font plus
\BIBentryALTinterwordstretchfactor\fontdimen3\font minus
  \fontdimen4\font\relax}
\providecommand{\BIBforeignlanguage}[2]{{%
\expandafter\ifx\csname l@#1\endcsname\relax
\typeout{** WARNING: IEEEtran.bst: No hyphenation pattern has been}%
\typeout{** loaded for the language `#1'. Using the pattern for}%
\typeout{** the default language instead.}%
\else
\language=\csname l@#1\endcsname
\fi
#2}}
\providecommand{\BIBdecl}{\relax}
\BIBdecl

\bibitem{menezes96crypto}
A.~J. Menezes, S.~A. Vanstone, and P.~C.~V. Oorschot, \emph{Handbook of Applied
  Cryptography}.\hskip 1em plus 0.5em minus 0.4em\relax CRC Press, Inc., 1996.

\bibitem{rubin93synth}
D.~B. Rubin, ``{Discussion of Statistical Disclosure Limitation},''
  \emph{{Journal of Offical Statistics}}, vol.~9, no.~2, pp. 461--468, 1993.

\bibitem{Dwork2008}
C.~Dwork, ``Differential privacy: A survey of results,'' in \emph{Proceedings
  of International Conference on Theory and Applications of Models of
  Computation}.\hskip 1em plus 0.5em minus 0.4em\relax Springer, 2008, pp.
  1--19.

\bibitem{jacob16opacity}
R.~Jacob, J.-J. Lesage, and J.-M. Faure, ``Overview of discrete event systems
  opacity: Models, validation, and quantification,'' \emph{Annual Reviews in
  Control}, vol.~41, pp. 135--146, 2016.

\bibitem{prorok2016macroscopic}
A.~Prorok and V.~Kumar, ``A macroscopic privacy model for heterogeneous robot
  swarms,'' in \emph{Proceedings International Conference on Swarm
  Intelligence}.\hskip 1em plus 0.5em minus 0.4em\relax Springer, 2016, pp.
  15--27.

\bibitem{OKa08}
J.~M. O'Kane, ``On the value of ignorance: Balancing tracking and privacy using
  a two-bit sensor,'' in \emph{Proceedings of International Workshop on the
  Algorithmic Foundations of Robotics}, 2008, pp. 235--249.

\bibitem{zhang18complete}
Y.~Zhang and D.~A. Shell, ``{Complete characterization of a class of
  privacy-preserving tracking problems},'' \emph{International Journal of
  Robotics Research---in WAFR'16 Special Issue}, 2018.

\bibitem{saberifar18pgraph}
F.~Z. Saberifar, S.~Ghasemlou, D.~A. Shell, and J.~M. O'Kane, ``{Toward a
  language-theoretic foundation for planning and filtering},''
  \emph{International Journal of Robotics Research---in WAFR'16 Special Issue},
  2018.

\bibitem{zhang18planning}
Y.~Zhang, D.~A. Shell, and J.~M. O'Kane, ``Finding plans subject to
  stipulations on what information they divulge,'' in \emph{Proceedings of
  International Workshop on the Algorithmic Foundations of Robotics}, 2018.

\end{thebibliography}
\end{document}